\documentclass{article}
\usepackage{arxiv}
\usepackage[utf8]{inputenc} 
\usepackage[T1]{fontenc}    
\usepackage{url}  
\usepackage{hyperref}       
\hypersetup{
    colorlinks=true,       
    linkcolor=blue,          
    citecolor=blue,        
    urlcolor=blue           
}
\usepackage{booktabs}       
\usepackage{nicefrac}       
\usepackage{microtype}      
\usepackage{lipsum}		
\usepackage{graphicx}
\usepackage{doi}
\usepackage{amsmath}
\usepackage{subfig}
\usepackage[affil-it]{authblk}
\usepackage[english]{babel}
\usepackage{blindtext}
\usepackage{amssymb}
\usepackage{cleveref}
\usepackage{booktabs}
\usepackage{framed,multirow}
\usepackage{hyperref}
\usepackage{url}            
\usepackage{booktabs}
\usepackage[mathlines]{lineno}
\usepackage{amsfonts}       
\usepackage{latexsym}
\usepackage{amsmath}
\usepackage{amssymb}
\usepackage{amsfonts}

\usepackage{cleveref}
\usepackage{nicefrac}       
\usepackage{microtype}      
\usepackage{lipsum}
\usepackage{graphicx}
\usepackage{adjustbox}
\usepackage{graphicx}
\usepackage[table,xcdraw]{xcolor}
\definecolor{newcolor}{rgb}{.8,.349,.1}
\usepackage[linesnumbered,ruled,vlined]{algorithm2e}

\usepackage{multirow}
\usepackage{diagbox}
\usepackage{subfig}
\usepackage{tikz}
\usetikzlibrary{external}

\usepackage{pgfplots}
\pgfplotsset{compat=1.7}
\usetikzlibrary{positioning}
\usepackage{etoolbox} 
\usepackage[numbers]{natbib}

\newcommand*\linenomathpatch[1]{%
  \cspreto{#1}{\linenomath}%
  \cspreto{#1*}{\linenomath}%
  \csappto{end#1}{\endlinenomath}%
  \csappto{end#1*}{\endlinenomath}%
}
\newcommand*\linenomathpatchAMS[1]{%
  \cspreto{#1}{\linenomathAMS}%
  \cspreto{#1*}{\linenomathAMS}%
  \csappto{end#1}{\endlinenomath}%
  \csappto{end#1*}{\endlinenomath}%
}
\expandafter\ifx\linenomath\linenomathWithnumbers
  \let\linenomathAMS\linenomathWithnumbers
  \patchcmd\linenomathAMS{\advance\postdisplaypenalty\linenopenalty}{}{}{}
\else
  \let\linenomathAMS\linenomathNonumbers
\fi

\linenomathpatch{equation}
\linenomathpatchAMS{gather}
\linenomathpatchAMS{multline}
\linenomathpatchAMS{align}
\linenomathpatchAMS{alignat}
\linenomathpatchAMS{flalign}

\SetCommentSty{mycommfont}

\crefformat{section}{\S#2#1#3}
\crefformat{subsection}{\S#2#1#3}
\crefformat{subsubsection}{\S#2#1#3}
\crefrangeformat{section}{\S\S#3#1#4 to~#5#2#6}
\crefmultiformat{section}{\S\S#2#1#3}{ and~#2#1#3}{, #2#1#3}{ and~#2#1#3}

\date{}
\title{Critical Investigation of Failure Modes in Physics-informed Neural Networks}

\author{\href{https://orcid.org/0000-0002-1095-0881}{\includegraphics[scale=0.08]{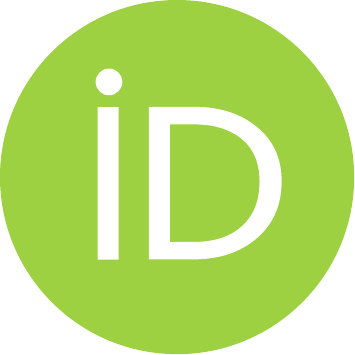}\hspace{1mm}Shamsulhaq Basir}\thanks{shb105@pitt.edu (Shamsulhaq Basir}}

\author{\href{https://orcid.org/0000-0003-1967-7583}{\includegraphics[scale=0.08]{orcid.pdf}\hspace{1mm}Inanc Senocak\thanks{corresponding author:~senocak@pitt.edu (Inanc Senocak)}}}

\affil{Department of Mechanical Engineering and Materials Science, University of Pittsburgh, \\ 3700 O'Hara St., Pittsburgh, PA 15261, USA}
\renewcommand{\shorttitle}{Critical Investigation of Failure Modes in Physics-informed Neural Networks}

\begin{document}
\maketitle
\begin{abstract}
Several recent works in scientific machine learning have revived interest in the application of neural networks to partial differential equations (PDEs). A popular approach is to aggregate the residual form of the governing PDE and its boundary conditions as soft penalties into a composite objective/loss function for training neural networks, which is commonly referred to as physics-informed neural networks (PINNs). In the present study, we visualize the loss landscapes and distributions of learned parameters and explain the ways this particular formulation of the objective function may hinder or even prevent convergence when dealing with challenging target solutions. We construct a purely data-driven loss function composed of both the boundary loss and the domain loss. Using this data-driven loss function and, separately, a physics-informed loss function, we then train two neural network models with the same architecture. We show that incomparable scales between boundary and domain loss terms are the culprit behind the poor performance. Additionally, we assess the performance of both approaches on two elliptic problems with increasingly complex target solutions. Based on our analysis of their loss landscapes and learned parameter distributions, we observe that a physics-informed neural network with a composite objective function formulation produces highly non-convex loss surfaces that are difficult to optimize and are more prone to the problem of vanishing gradients.
\end{abstract}
\section{Introduction}
Deep learning has shown tremendous success in the pattern recognition \cite{krizhevsky2012imagenet,he2016deep}, speech recognition \cite{hinton2012deep}, and natural language processing \cite{sutskever2014sequence,weston2014tagspace,chowdhury2003natural}. The success of these models exudes collectively from the fast development of accessible information, increment in computing power, and advanced learning algorithms that have made strides in the understanding of neural networks \cite{schmidhuber2015deep}.
The introduction of the universal approximation theorem \cite{hornik1989multilayer,leshno1993multilayer} has stimulated new studies using neural networks to solve ODEs and PDEs. \citet{dissanayake1994neural} pioneered the use of neural networks to solve PDEs, where they assembled the residual form of a given PDE and its boundary conditions as soft-constraints for training their neural network model. \citet{van1995neural} presented a similar approach and showed its potential on a  magnetohydrodynamics plasma equilibrium problem. The general neural network-based approach that was proposed in \cite{van1995neural,dissanayake1994neural} was applied with satisfactory outcomes to the non-linear Schrodinger equation in \cite{monterola2001solving}, to a non-steady fixed bed non-catalytic solid/gas reactor problems in \cite{parisi2003solving}, and to the one-dimensional Burgers equation in \cite{hayati2007feedforward}. \citet{lagaris1998artificial} proposed a neural network-based approach for the solution of ODEs and PDEs on orthogonal box domains by forming trial functions that satisfy the boundary conditions by construction. It is worth mentioning that these early works did not find broader adoption or appreciation by other researchers likely because of a lack of computational resources and an incomplete understanding of neural networks at the time of their introduction. Modern machine learning frameworks with automatic differentiation capabilities \cite{abadi2016tensorflow,paszke2019pytorch} have revived the use of neural networks to solve ODEs and PDEs. Several recent investigations \cite{yu2017deep,Raissi2019,sirignano2018dgm} adopted the overall approach and formulation proposed in \cite{dissanayake1994neural,van1995neural}. \citet{Raissi2019} coined the term physics-informed neural networks (PINNs) to describe this technique, which has sparked considerable interest in the application of neural networks for the solution of problems involving PDEs and ODEs. 

In this work, we investigate the failure modes of the PINN approach in the context of elliptic PDEs. We formulate an equivalent but purely data-driven objective function composed of the boundary loss and the domain loss term to contrast it against the loss function formulation adopted in the PINN approach. 
We then train two neural network models of the same architecture with the physics-informed and the purely data-driven objective functions and investigate their performances by visualizing their loss landscapes and the distribution of their learned parameters. The paper is organized as follows. Section III presents the technical background on supervised and unsupervised learning approaches. In section IV we compare three objective/loss formulations, one of which is the recently introduced \emph{Physics and Equality Constrained Artificial Neural Networks}(PECANNs) \cite{basir2021physics} through numerical experiments on two elliptic PDE problems and provide insights from these comparisons. Finally, section V concludes the paper with a discussion of our findings and future directions.
\section{Technical Background}\label{sec:tech_back}
\begin{figure}[!ht]
    \centering
   \includegraphics[width=0.65\textwidth]{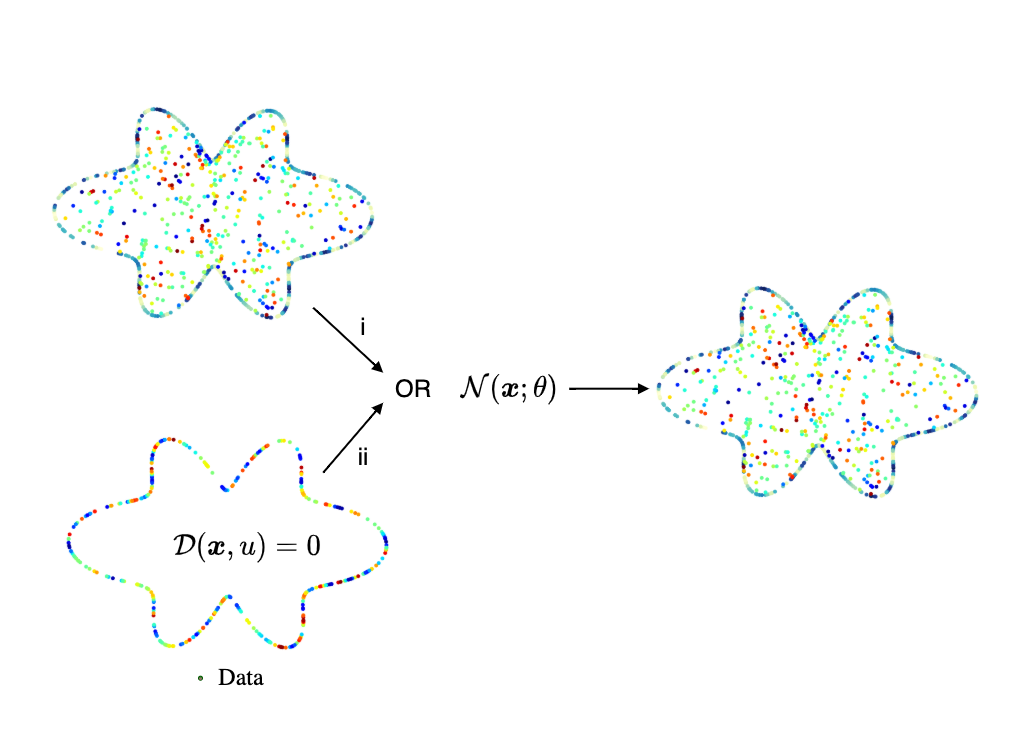}
    \caption{Schematic representation of learning approaches with an artificial neural network model $\mathcal{N}$ with its inputs $\boldsymbol{x}$ and its parameters $\theta$: the \textit{i} branch represent supervised learning approach, whereas the \textit{ii} branch represents physics-based unsupervised learning approach}
    \label{fig:training_procedure}
\end{figure}

In this section, we start with describing supervised regression for function approximation using data from the exact solution of the PDE alone. We then describe physics informed/constrained neural networks for the solution of PDEs. Let us consider a scalar function $u(\boldsymbol{x}):\mathbb{R}^d \rightarrow \mathbb{R}$ satisfying the following differential equation
\begin{align}
    \mathcal{D}(\boldsymbol{x},u) &:= 0 ~\text{in} ~\Omega, \label{eq:governing_PDE}\\
    u(\boldsymbol{x}) &:= g(\boldsymbol{x}) ~\text{on} ~\partial \Omega,\label{eq:governing_PDE_BC}
\end{align}
where $\Omega$ is the domain and $\partial \Omega$ is its boundary. We aim to learn a parameterized solution $u_{\theta}(x)$ represented by a neural network model that satisfies the governing equation and its boundary condition as in Eq.~\eqref{eq:governing_PDE}. We consider two different learning approaches. In the first approach, the solution $u_{\theta}$ is learned in a supervised fashion should there exist enough training data. For our investigation, we derive the training data from the exact solution to the PDE. In the second approach, the underlying physics through the governing PDE can be accounted for in the objective function for training a neural network model to learn $u_{\theta}$ in an unsupervised fashion. A schematic representation of both learning approaches are depicted in Fig~\ref{fig:training_procedure}.
\subsection{Supervised Learning Method}\label{sec:SupervisedRegression}
In this section, we describe a purely data-driven learning approach. We aim to have a composite loss function with no scale disparity between the domain loss and the boundary loss. To this end, we assume that there are sufficient training data on the boundary and in the domain. We describe a purely data-based objective formulation as a sum of mean-squared-errors (MSE) of loss terms in the domain and on the boundaries as in Eq.~\eqref{eq:supervisedLearningLoss} 
\begin{align}
    \mathcal{L}_{\Omega} ~&= \frac{1}{N_{\Omega}}\sum_{i=1}^{N_{\Omega}}\big[u_{\theta}(\boldsymbol{x}^{(i)}) - u(\boldsymbol{x}^{(i)}) \big]^2,\\
    \mathcal{L}_{\partial \Omega} &= \frac{1}{N_{\partial \Omega}}\sum_{j=1}^{N_{\partial \Omega}}\big[u_{\theta}(\boldsymbol{x}^{(j)}) - g(\boldsymbol{x}^{(j)})\big]^2,\\ 
     \mathcal{L}~~~&= ~ \mathcal{L}_{\Omega} + \mathcal{L}_{\partial \Omega},
    \label{eq:supervisedLearningLoss}
\end{align}
where $\mathcal{L}$ is the objective function,  $N_{\partial \Omega}$ and $N_{\Omega}$ are the number of training points on the boundary and in the domain respectively. 
\subsection{Unsupervised Learning Methods}
In this section, we present the popular physics-informed neural network method and a newly introduced, alternative method based on a constrained optimization formulation of the artificial neural networks \citet{basir2021physics}.
\subsubsection{PINN: Physics-informed Neural Networks }\label{sec:PINNs}
Here, we describe the common elements of the physics-informed learning framework presented in the works of \citet{dissanayake1994neural}, \citet{van1995neural}, and \citet{raissi2017physicsII}. In all these works, the domain and boundary loss terms are aggregated together as soft penalties. 
\begin{align}
    \mathcal{L}_{\Omega} ~&= \frac{1}{N_{\Omega}}\sum_{i=1}^{N_{\Omega}} \big[\mathcal{D}(\boldsymbol{x}^{(i)}, u_{\theta}(\boldsymbol{x}^{(i)}) \big]^2,\\
    \mathcal{L}_{\partial \Omega} &= \frac{1}{N_{\partial \Omega}}\sum_{j=1}^{N_{\partial \Omega}} \big[ g(\boldsymbol{x}^{(j)}) - u_{\theta}(\boldsymbol{x}^{(j)}) \big]^2,\\
    \mathcal{L}~~~&= ~\mathcal{L}_{\Omega} + \mathcal{L}_{\partial \Omega} 
    \label{eq:unsupervisedLearningLoss}
\end{align}
Note that each loss term is scaled or weighted by the number of training points. The aggregate loss $\mathcal{L}$ constitutes a composite objective function. 

\subsubsection{PECANN: Physics \& Equality Constrained Artificial Neural Networks}\label{sec:PECANNs}
In this section, we describe a recent physics-based unsupervised learning approach that combines a constrained optimization approach with modern deep learning practices for the solution of problems involving PDEs and ODEs \cite{basir2021physics}. The core idea is to minimize a domain loss based on the residual form of Eq.~\eqref{eq:governing_PDE} such that it is constrained by the boundary conditions (i.e. Eq.~\ref{eq:governing_PDE_BC}). This formally constrained optimization formulation is then put into practice for training an artificial neural network architecture using the Augmented Lagrangian method (ALM) \cite{powell1969method,hestenes1969multiplier}.  
\begin{align}
    \mathcal{L}_{\Omega} ~&=\sum_{i=1}^{N_{\Omega}} \big[\mathcal{D}(\boldsymbol{x}^{(i)}, u_{\theta}(\boldsymbol{x}^{(i)}) \big]^2\\
    \mathcal{L}_{\mu} &= \mathcal{L}_{\Omega} + \boldsymbol{\lambda}^T \mathcal{C} + \frac{\mu}{2}\|\mathcal{C}\|^2,\\
    \boldsymbol{\lambda} ~~~&=\boldsymbol{\lambda} + \mu \mathcal{C},
\end{align}
where $\mu$ is a positive penalty parameter with a maximum safeguarding value of $\mu_{\infty}$, $\boldsymbol{\lambda}^T =\{\lambda^{(1)}, \cdots,\lambda^{(N_{\partial \Omega})} \} $ is a vector of Lagrange multipliers, $\mathcal{C}^T = \{[g(\boldsymbol{x}^{(1)}) - u_{\theta}(\boldsymbol{x}^{(1)})]^2,\cdots, [g(\boldsymbol{x}^{(N_{\partial \Omega})}) - u_{\theta}(\boldsymbol{x}^{(N_{\partial \Omega})})]^2 \}$ is a vector of evaluated boundary constraints and $T$ is transposition.
\section{Numerical Experiments}
In this section, we present an in-depth investigation of how supervised and unsupervised neural networks perform through loss landscape visualizations and distributions of learned parameters. We target elliptic problems as they are central to the numerical solution of incompressible flows. 
\subsection{Simulation Parameters}
We use a feed-forward neural network with eight hidden layers and 20 neurons per layer with $\tanh$ activation functions. We use Adam \cite{kingma2017adam} with $10^{-2}$ initial learning rate and ReduceLROnPlateau(patience=100,factor=0.90) as the learning rate scheduler. Given two sets of vectors $\boldsymbol{\hat{u}}$, the learned solution, and $\boldsymbol{u}$, the exact solution, we define the relative $L_{2,r}$ of $\boldsymbol{\hat{u}}$ as follows,
\begin{equation}
     L_{2,r} = \frac{\|\hat{\boldsymbol{u}} - \boldsymbol{u}\|_2}{\|\boldsymbol{u}\|_2}.
    \label{eq:relativeL2Error}
\end{equation}
\subsection{One Dimensional Poisson's Equation}\label{sec:Poisson_1D}
\begin{align}
    \frac{d^2 u(x)}{dx^2} &= f(x), ~\text{in} ~ \Omega, \\
    u(x) &= g(x), ~ \text{on} ~\partial \Omega,
\label{eq:poissonOneDim}
\end{align}
where $f(x)$ and $g(x)$ are forcing functions, $\Omega = \{x~|~ 0 \ge x \le 1\}$ and $\partial \Omega$ is its boundary. We manufacture a solution of the form $u(x; \omega) = \sin(\omega \pi x)$ and use it to calculate $f(x)$ and $g(x)$. The particular form the manufactured solution is desirable as we can increase its complexity simply by increasing $\omega$. For the purpose of supervised learning, we generate 600 training data in the domain and two boundary data from the boundary conditions and simply fit the data and learn the function. For the purpose of physics based learning, we generate 600 collocation points that we use to calculate on the residual form of Eq.~\eqref{eq:poissonOneDim} and two boundary points with their respective boundary values from the exact equation at every epoch.
\subsubsection{Predictions}
\begin{figure}[!ht]
\centering
\includegraphics[width=0.85\textwidth]{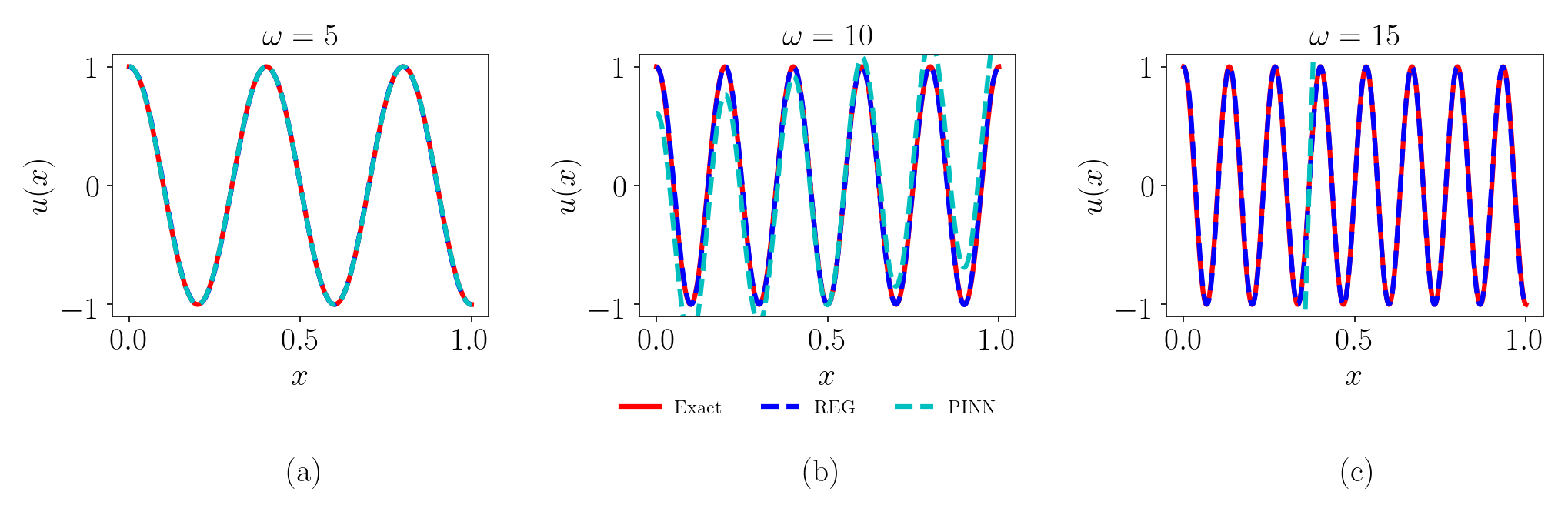}
\caption{\textit{One dimensional Poisson's equation: predicted solution $\hat{u}(x)$ from PINN model (dashed cyan) and purely data-driven regression model (dashed blue)} (a) wave number $\omega = 5$ with $L_{2,r} = 1.028\times 10^{-2}$ and $L_{2,r} = 8.269\times 10^{-4}$ for PINN and regression model respectively, (b) wave number $\omega = 10$ with $L_{2,r} = 2.087\times 10^{-2}$ and $L_{2,r} = 3.195\times 10^{-1}$ for PINN and regression model respectively, (c) wave number $\omega = 15$ with $L_{2,r} = 3.077 \times 10^{-2}$ and $L_{2,r} = 1.652\times 10^{+1}$ for PINN and regression model respectively}
 \label{fig:OneDimPoisson}
\end{figure}

In this section, we present predictions for increasingly complex functions from the neural network models trained using supervised and unsupervised learning approaches. The results of our experiment is presented in Fig~\ref{fig:OneDimPoisson}. 

We observe that as we increase the complexity of the target function, the neural network model trained with a supervised regression approach converges under fixed hyperparameter settings. However, the same behavior is not observed from the same neural network model trained using PINN learning approach. Thus far, we know that for the supervised learning approach, both terms in the loss function have a similar scale without any derivative terms as seen in Eq.~\eqref{eq:supervisedLearningLoss}. However, for the physics-informed learning approach, we have a disparity of scales between each loss term since the calculation of the residual form of the PDE in Eq. \eqref{eq:governing_PDE} involves derivative terms. Therefore, we observe that aggregating two terms with different scales into a single objective function works for a fairly simple target solution and fails or produces erroneous results with challenging target functions. By this experiment, we showed that this failure is not due to inadequacy of expressivity in the neural network architecture but rather due to the formulation of physics-informed objective/loss function.

\subsubsection{Investigations}
\begin{figure}[!ht]
\centering
\subfloat{\includegraphics[width=0.85\textwidth]{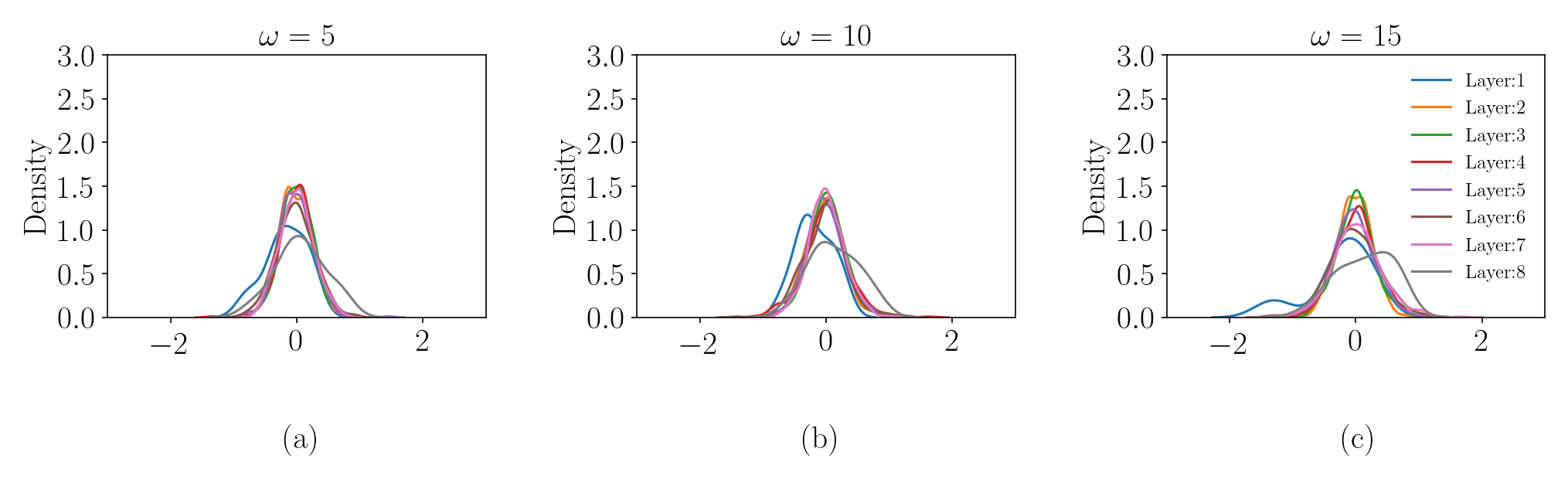}}\\
\subfloat{\includegraphics[width=0.85\textwidth]{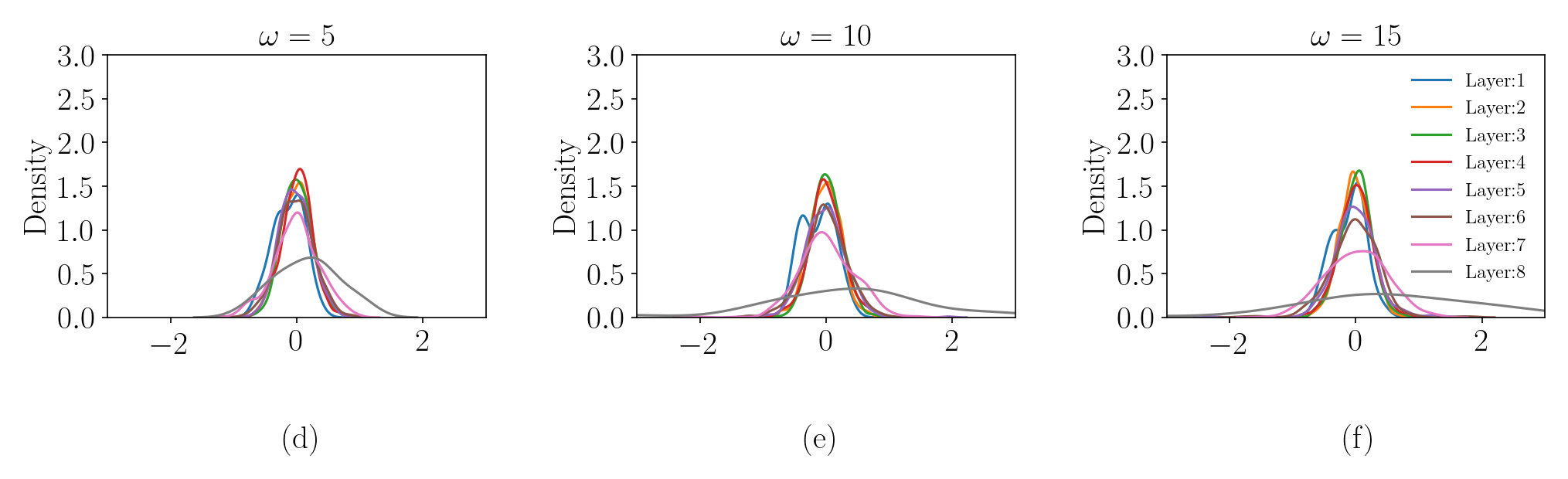}}
\caption{\textit{Weight histograms of neural network models trained using two different approaches} Top row: using supervised learning approach. (a) wave number $\omega = 5$, (b) wave number $\omega = 10$ , (c) wave number $\omega = 15$. Bottom row: using PINN approach. (d) wave number $\omega = 5$ , (e) wave number $\omega = 10$, (f) wave number $\omega = 15$.}
 \label{fig:oneDimPoissonWeightKDE}
\end{figure}

We now investigate the above failure mode by comparing the histograms of learned parameters by both models as well as their loss surfaces.  In Fig.~\ref{fig:oneDimPoissonWeightKDE}, we present histograms of learned parameters by the purely data-driven supervised neural-network model and the unsupervised PINN model. Our aim here is to investigate the differences between the histograms of parameters in both models. It is worth noting that we initialize all the parameters in the network with Xavier initialization technique\cite{glorot2010understanding}.

\begin{figure}[!ht]
\centering
\subfloat{\includegraphics[width=0.85\textwidth]{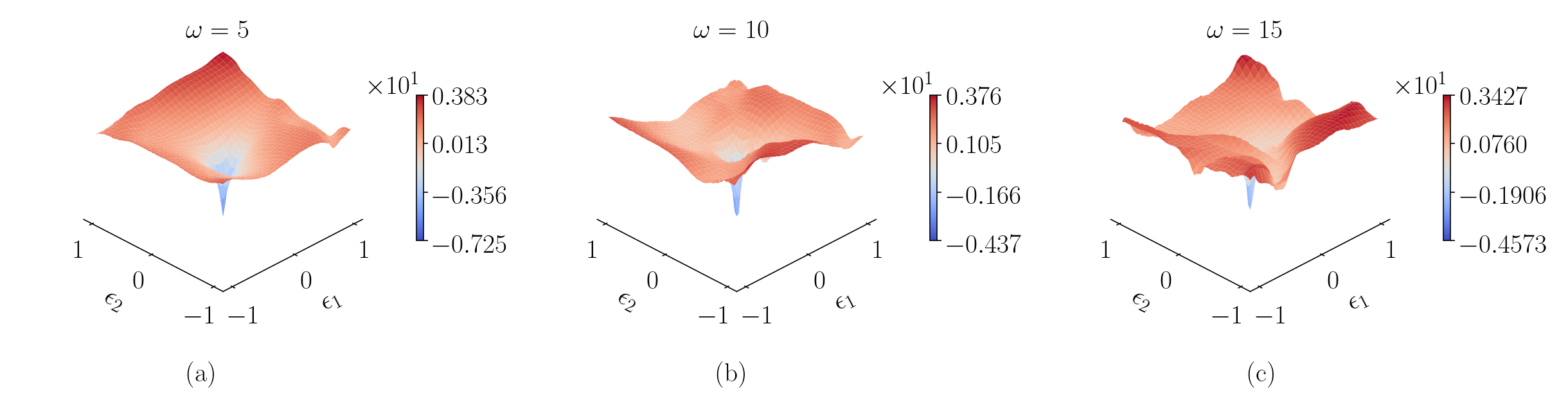}}\\
\subfloat{\includegraphics[width=0.85\textwidth]{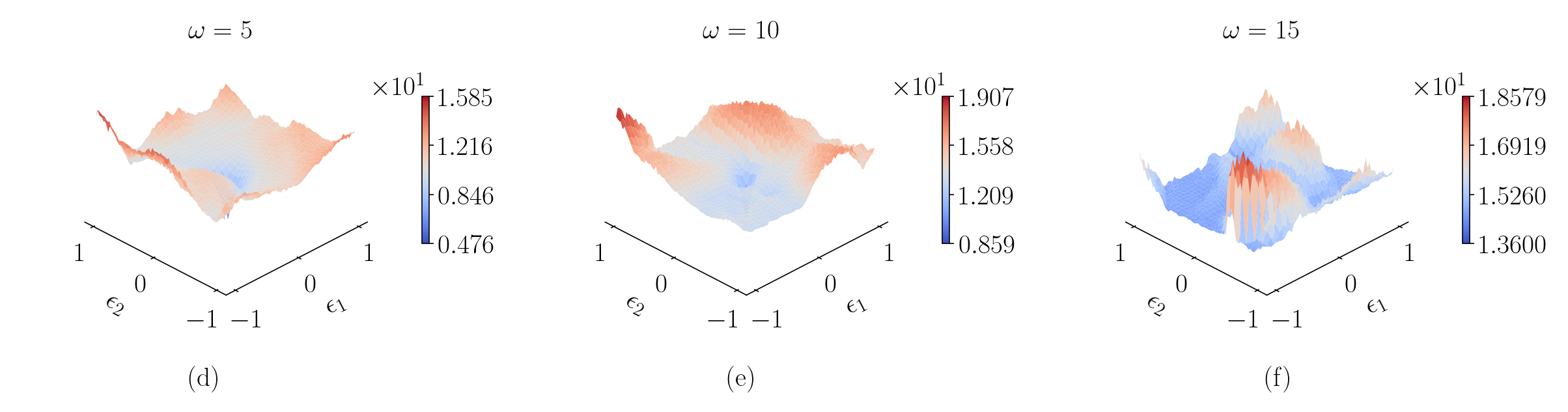}}
\caption{\textit{Loss landscape of the neural network model.} Top row: purely-data driven regression model. (a) wave number $\omega = 5$, (b) wave number $\omega = 10$ , (c) wave number $\omega = 15$. Bottom row: physics-informed neural network model. (d) wave number $\omega = 5$ , (e) wave number $\omega = 10$, (f) wave number $\omega = 15$.}
 \label{fig:oneDimPoissonLossLandscape}
\end{figure}

From Figs.~\ref{fig:oneDimPoissonWeightKDE}(a-c), we observe that as we increase the complexity of the target function (i.e. larger $\omega$), weight histograms of the neural network model trained using supervised learning approach stay fairly the same for all the cases. 
However, increasing the complexity of the target function (i.e. larger $\omega$) changes the distribution of learned weights for the unsupervised PINN model as seen in Figs.~\ref{fig:oneDimPoissonWeightKDE}(d-f). 
We attribute this observation for physics-informed neural network to the derivatives in the governing equation and conclude that learning becomes harder for physics-informed neural networks when the target problem gets complex. We further investigate the loss landscapes by perturbing the trained models across two random directions \cite{li2017visualizing}. We present loss surfaces extracted from the neural-network models trained using both learning approaches in Fig.~\ref{fig:oneDimPoissonLossLandscape}
We observe that as we increase the complexity of target function from $\omega=5$ to $\omega=15$, loss surfaces of the neural network model trained using supervised learning approach behave well with a visible minimum as seen in Figs.~\ref{fig:oneDimPoissonLossLandscape}(a)-(c).
However, with the increasing complexity of the target function, loss surfaces the PINN model becomes complex and hard to optimize as seen in Fig.~\ref{fig:oneDimPoissonLossLandscape}(d)-(f).
These experiments show that for physics-informed neural networks loss surfaces become complex with the complexity of the target function which is not desirable. Because learning complex functions become challenging as in the case of $\omega = 15$.

\subsubsection{Physics and Equality Constrained Artificial Neural Networks}
\begin{figure}[!ht]
\centering
\includegraphics[width=0.85\textwidth]{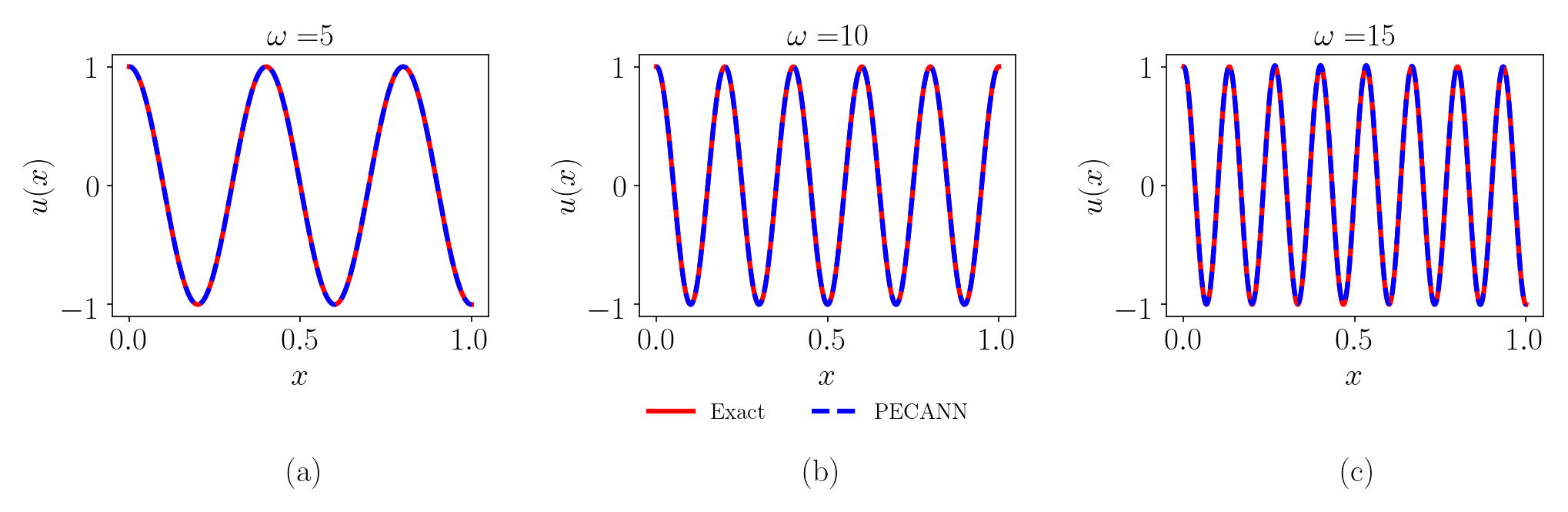}
\caption{\textit{One dimensional Poisson's equation: predicted solution $\hat{u}(x)$ from PECANN model (dashed cyan) versus exact solution} (a) wave number $\omega = 5$ with $L_{2,r} = 1.060 \times 10^{-4}$ , (b) wave number $\omega = 10$ with $L_{2,r} = 6.333 \times 10^{-4}$, (c) wave number $\omega = 15$ with $L_{2,r} = 9.627 \times 10^{-3}$}
 \label{fig:OneDimPoissonPECANN}
\end{figure}

Through numerical experiments, we have observed that it is possible to improve the predictions with the PINN model and its composite objective function formulation by manually tuning the weight of the boundary loss term. However, the composite objective function with user-defined weights is an inchoate optimization problem formulation, to begin with, and it is at the root of several performance issues associated with the PINN model.
In \citet{basir2021physics}, the authors pursued a constrained-optimization formulation and proposed the Physics\& Equality Constrained Artificial Neural Networks (PECANNs). In this framework, the authors use the Augmented Lagrangian method (ALM) \cite{powell1969method, hestenes1969multiplier} to convert a constrained-optimization problem to an unconstrained optimization problem suitable for neural networks. In this work, we show that training the same neural network architecture with the PECANN approach easily captures the target function with increasing difficulty. We present the predictions from our PECANN model and the exact solution in Fig.~\ref{fig:OneDimPoissonPECANN}.

\begin{figure}[!ht]
\centering
\includegraphics[width=0.85\textwidth]{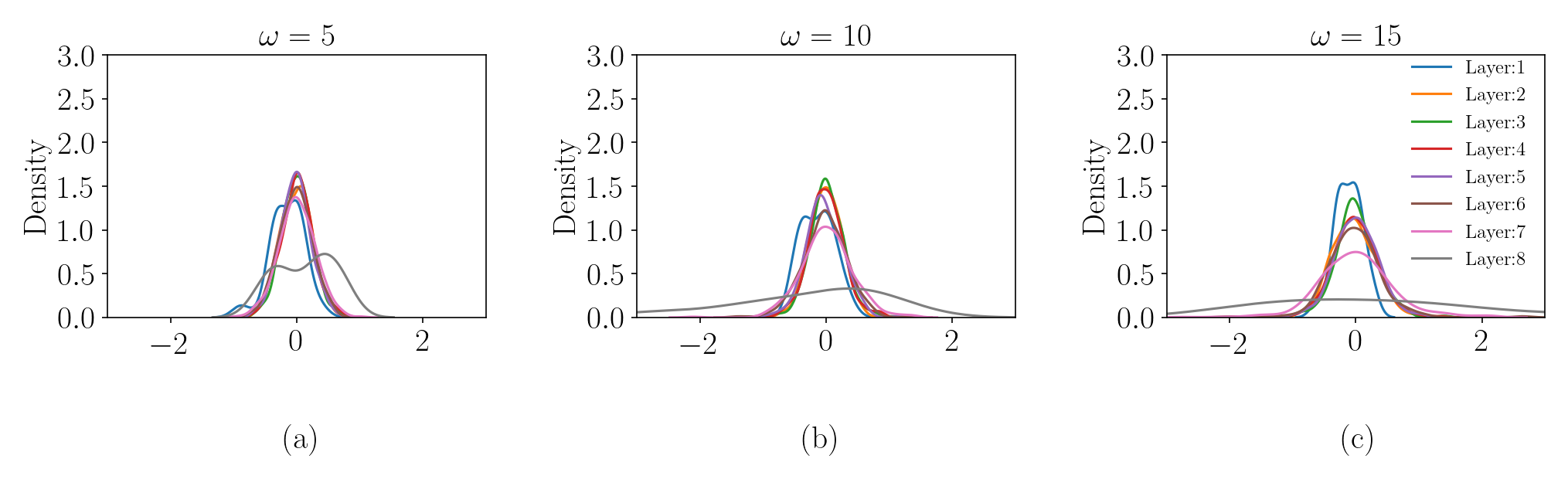}
\caption{\textit{Weight histograms of the PECANN model for one dimensional Poisson's equation.} (a) wave number $\omega = 5$ , (b) wave number $\omega = 10$ , (c) wave number $\omega = 15$ .}
 \label{fig:oneDimPoissonWeightKDEPECANN}
\end{figure}

We further investigate our trained model by presenting the histogram of its parameters for each layer in Fig.~\ref{fig:oneDimPoissonWeightKDEPECANN}. 

\begin{figure}[!ht]
\centering
\includegraphics[width=0.85\textwidth]{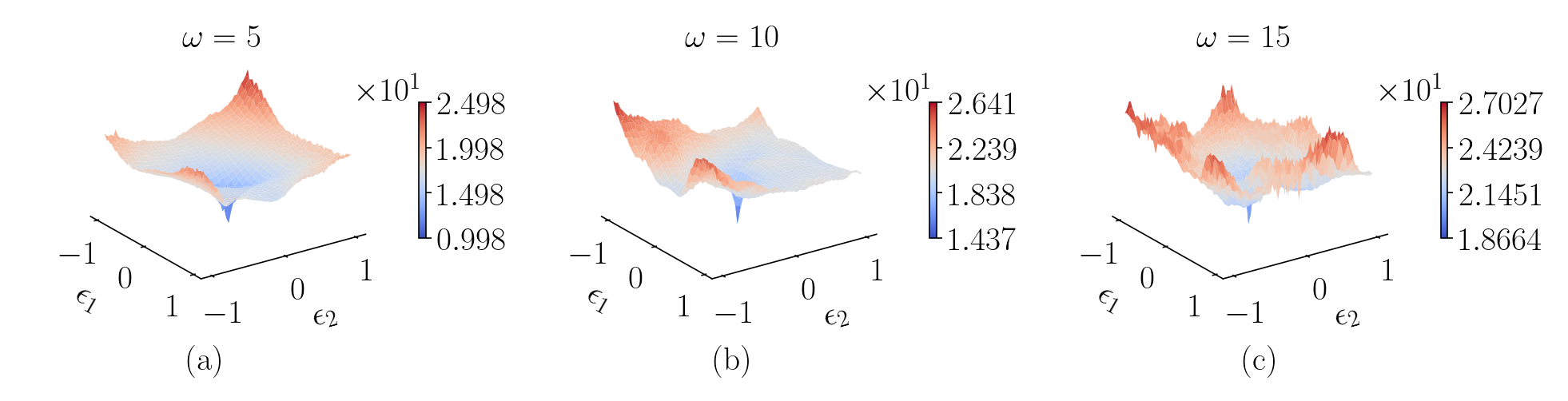}
\caption{\textit{Loss landscape of PECANN model for one dimensional Poisson's equation.} (a) wave number $\omega = 5$, (b) wave number $\omega = 10$ , (c) wave number $\omega = 15$}
 \label{fig:oneDimPoissonLossLandscapePECANN}
\end{figure}

From Fig.~\ref{fig:oneDimPoissonWeightKDEPECANN} we observe that as we increase the complexity of the target function, the histograms of weights change as was the case with the PINN model. This shift of weight distribution between hidden layers as we increase the complexity of the target function is due to structured physics in the objective function. In \cite{basir2021physics}, authors propose a simple residual neural network that has no extra weight or activation function compared with a vanilla neural network that addresses this issue and significantly speeds up training. For further insight, we present loss surfaces for three target functions in Fig.~\ref{fig:oneDimPoissonLossLandscapePECANN}. 
From Fig.~\ref{fig:oneDimPoissonLossLandscapePECANN}(a-c) we observe that loss landscapes stay smooth with a clear local minimum that the optimizer has found. Unlike the PINN model, our PECANN model has smooth loss landscapes that are easier to optimize. This is attributed to the unconstrained optimization formulation of the objective function that properly constrains the boundary conditions and addresses the scale discrepancy. 

\subsection{Two Dimensional Poisson's Equation}\label{sec:Poisson_2D}
In the previous numerical experiment, we have investigated the effect of scale by comparing a supervised learning approach, in which the domain and the boundary loss terms had the same scale, against an unsupervised learning approach where the domain loss term had derivative terms whereas its boundary loss term did not because of the imposed Dirichlet boundary conditions. We should note that the mean squared error is averaged of the number of training points for the domain and the boundary loss terms. Because we have only two boundary points in a one-dimensional problem, this exercise was not amenable to an in-depth investigation. Therefore, in this section, we consider a two-dimensional problem so that average loss values calculated for the boundary and the domain can be computed over the equal number of points. Hence, the purely data-driven supervised neural network model not only has the same scale for the domain and the boundary loss terms but is also averaged over equal number of points in both of the loss terms. Similarly, for the physics-informed neural network model, we have the equal number of points for both the domain and the boundary loss terms, but they end up having different scales because the domain loss term includes derivatives, whereas the boundary loss term does not because of the Dirichlet type conditions. This aspect of the problem enables us to observe the effect of structures or derivative terms in the domain loss in the physics-informed neural-network model. As for our PECANN model, which is also an unsupervised model, we do not have the issue of scaling anymore as it is taken care of by the augmented Lagrange method. Also regardless of where our PECANN model converges in the space of solutions, we make sure the obtained solutions satisfy the constraints or in other words, the solution is feasible. 

We consider the following two-dimensional non-homogeneous Poisson's equation:
\begin{align}
\nabla^2 u(x,y) &= f(x,y),~(x,y) ~\text{in}~\Omega,\\
u(x,y) &=g(x,y),~(x,y)~\text{on} ~ \partial \Omega,
\label{eq:2D_Poisson}
\end{align}
where $\Omega = \{(x,y) ~|~ 0 \leq x \leq 1,~0 \leq y \leq 1\}$, $f$ and $g$ are source functions across the domain and the boundary respectively.
We manufacture a solution as follows
\begin{align}
        u(x,y)  = \cos(15 \pi x) \exp( - \pi y),\qquad \forall (x,y) ~\text{in}~\Omega.
        \label{eq:2D_Poisson_exact}
\end{align}
In the case of supervised learning, we generate 600 training data in the domain and 600 boundary data from the Dirichlet boundary conditions and learn the function by simply fitting the data with a neural network. In the case of unsupervised, physics-based learning, we generate 600 collocation points that we use to calculate the residual form of Eq.~\eqref{eq:2D_Poisson} and 600 boundary points with their respective boundary values from the exact equation Eq.~\eqref{eq:2D_Poisson_exact} at every epoch. We use the same network architecture, optimizer, and learning rate scheduler as in the previous problem. We set $\mu_{\infty} = 500$ in the PECANN model. We present the predicted solutions from three neural network models trained with the purely data-driven supervised learning approach and the unsupervised learning approaches with PINN and PECANN. 
\begin{figure}[!h]
    \centering
    \includegraphics[width=\textwidth]{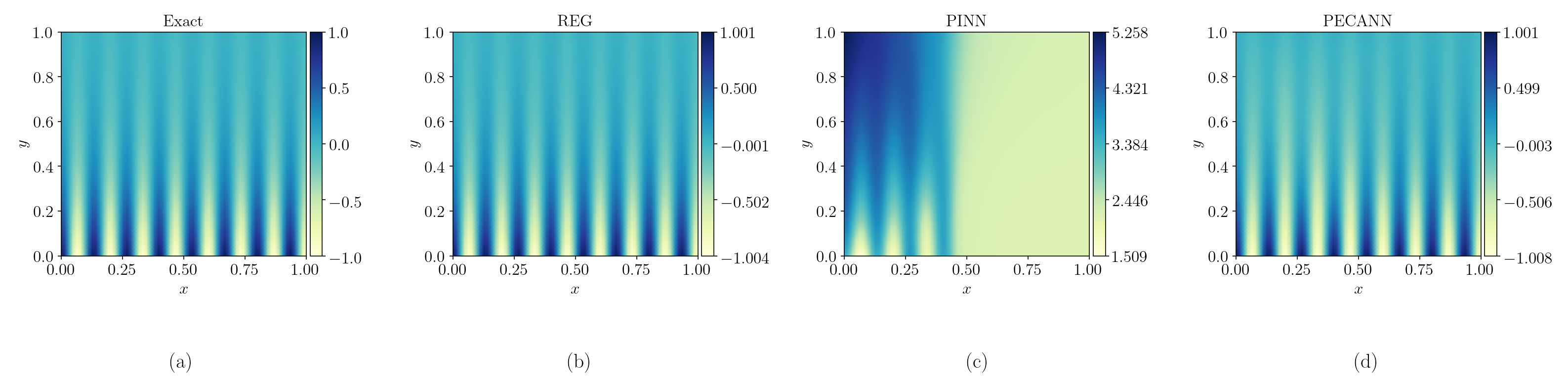}
    \caption{Two dimensional Poisson's equation: (a) exact solution $u(x,y)$, (b) predicted solution from purely data-driven supervised neural network model, (c) predicted solution from unsupervised physics-informed neural network model(PINN), (d) predicted solution from unsupervised physics-and-equality constrained artificial neural network model (PECANN)}
    \label{fig:poisson_two_dim}
\end{figure}

From Fig.~\ref{fig:poisson_two_dim}(b), we observe that the purely data-driven supervised neural network model has converged to the exact solution which implies that our neural network architecture is capable of representing the solution function. However, Fig.~\ref{fig:poisson_two_dim}(c), shows that the unsupervised physics-informed neural network model has diverged. We note that the only major difference between the supervised model and the PINN model is the loss term for the domain. Therefore, we see that lumping two loss terms with different scales impedes the convergence of neural network models. However, unlike the PINN model, we observe from Fig.~\ref{fig:poisson_two_dim}(d) that the unsupervised PECANN model has converged to the correct solution. In the PECANN model, the difference in the scale of the domain and boundary loss terms are taken care of by the augmented Lagrangian method which gets updated during training to make sure the network converges to a minimum such that the constraints are satisfied. 
\begin{figure}[!ht]
    \centering
    \includegraphics[width=\textwidth]{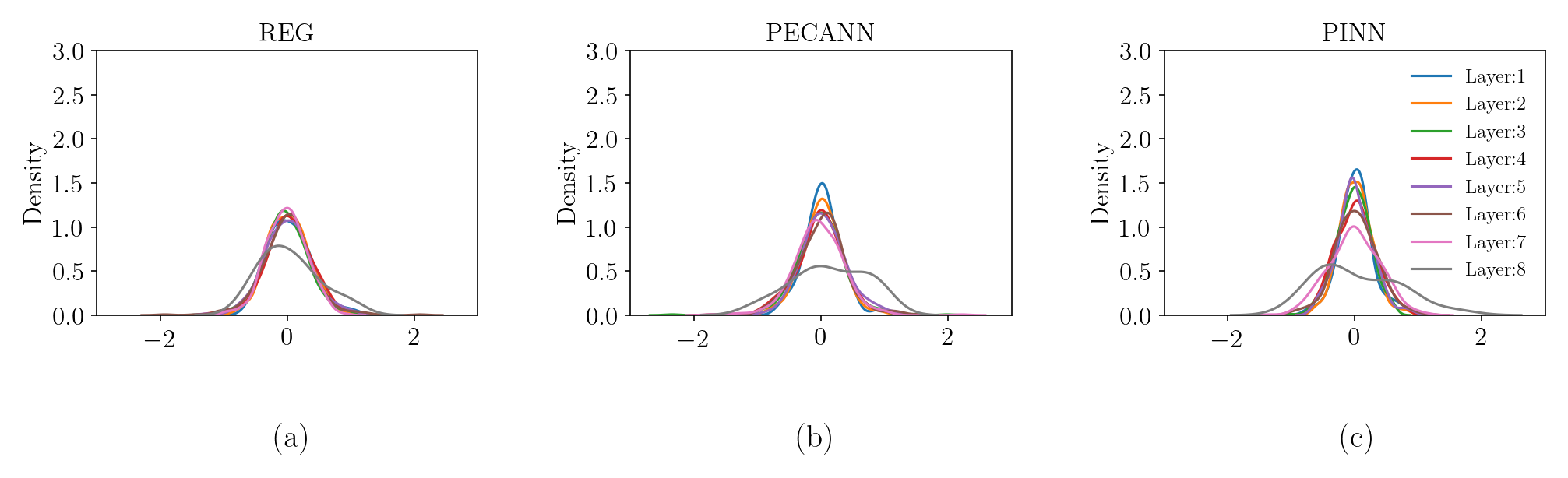}
    \caption{Weight histograms of neural network models trained for the solution of a two dimensional Poisson's equation: (a) purely data-driven supervised neural network model, (b) unsupervised physics-informed neural network model, (c) unsupervised physics-and-equality constrained artificial neural network model}
    \label{fig:poisson_two_dim_weight_histograms}
\end{figure}

In Fig.~\ref{fig:poisson_two_dim_weight_histograms}, we present histograms of the learned weights for all the neural network models considered for the two-dimensional Poisson's equation problem. From Fig.~\ref{fig:poisson_two_dim_weight_histograms}(a), we observe that layers 1 through 7 have similar means and variances. However, having a slightly higher peak in their distribution than layer 8 indicates that they must have suffered from a mild vanishing gradient problem. However, Fig.~\ref{fig:poisson_two_dim_weight_histograms}(b -c), show that embedding the physics into the neural network slightly exacerbates the problem of vanishing gradient regardless of the scale difference between the loss terms. Because PECANN model does not have the issue of scale disparity but the parameter distribution of PECANN and PINN do not have any significant difference. This is a crucial observation and is because during training, predictions from the network are naturally noisy and their derivatives end up being noisy as well. Therefore, the neural network model is more vulnerable to the issue of vanishing gradients. As mentioned earlier this issue was addressed in \cite{basir2021physics} with their proposed residual neural network architecture. So far, we have shown that aggregating multiple loss terms without properly balancing them impedes the convergence of the neural network model and can cause the network to suffer from the vanishing gradient problem. At this point, we visualize the loss landscapes of these three models to gain more insight. 
\begin{figure}[!ht]
    \centering
    \includegraphics[width=\textwidth]{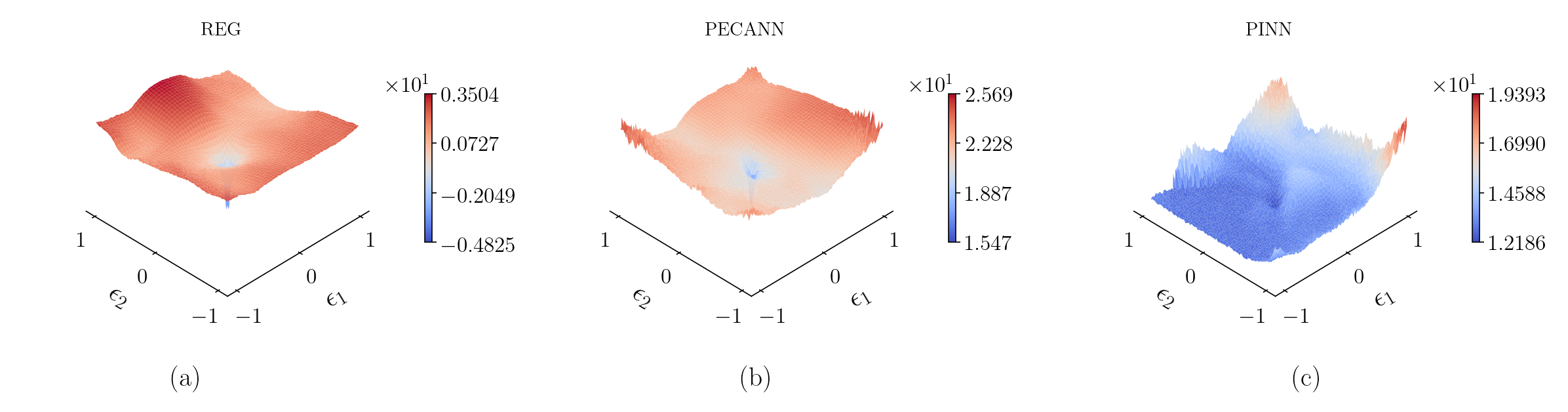}
    \caption{Loss landscapes of neural network models trained for the solution of a two dimensional Poisson's equation: (a) purely data-driven neural network model, (b) physics-informed neural network model, (c) physics-and-equality constrained artificial neural network model}
    \label{fig:poisson_two_dim_loss_landscapes}
\end{figure}
From Fig.~\ref{fig:poisson_two_dim_loss_landscapes}(a), we observe the loss landscape of the purely data-driven supervised neural network model is smooth with a clear minimum that the optimizer was able to find. 
In Fig.~\ref{fig:poisson_two_dim_loss_landscapes}(b), we observe that the PECANN model also has a smooth loss landscape with a clear minimum that is found by the optimizer. However, Fig.~\ref{fig:poisson_two_dim_loss_landscapes}(b) shows the loss landscape for the PINN model is highly non-convex with flat regions. Therefore, the optimizer was not able to get out and find a better local minimum.
\section{Conclusions}
We have investigated a particular network-based approach that aggregates the residual form of the PDE of interest and its boundary conditions into a composite objective function. We demonstrated that this particular formulation can severely limit the predictive power of these networks. Our analysis on the solution of elliptic problems revealed that PINN may work for simple target solutions and but fail to converge for more complex problems. We showed that this failure mode is due to the difference of scales between the boundary and the domain losses. For further insight into how the disparity of scales impact the loss landscapes of the neural network models, we visualized the histograms of their learned parameters and their loss landscapes. We observed that embedding physics into the loss function makes learning hard and causes the parameter distributions to shift in between layers. We observed that for the PINN model, loss surfaces become increasingly complex and hard to optimize for increasingly complex target functions unlike the loss surfaces for the purely data-driven supervised model that stays fairly smooth and easy to optimize. We then demonstrated that this underlying problem can be solved by properly formulating the objective function. We showed that a constrained optimization formulation of a well-posed PDE employing the Augmented Lagrange method successfully learns the target solutions for all the cases.
\section*{Acknowledgments}
This material is based upon work supported by the National Science Foundation under Grant No. (1953204). Research was sponsored in part by the University of Pittsburgh, Center for Research Computing through the computing resources provided.
\clearpage
\bibliographystyle{elsarticle-num-names} 
\bibliography{sample}
\end{document}